\documentclass[11pt,a4paper]{article}

\usepackage[margin=2.3cm]{geometry}
\usepackage[T1]{fontenc}
\usepackage[utf8]{inputenc}
\usepackage{amsmath,amssymb,amsfonts}
\usepackage{graphicx}
\usepackage{booktabs}
\usepackage{multirow}
\usepackage{tabularx}
\usepackage{hyperref}
\usepackage{enumitem}
\usepackage{caption}
\usepackage{subcaption}
\usepackage{xcolor}
\usepackage{longtable}
\usepackage{array}
\usepackage{natbib}
\usepackage{url}
\usepackage{setspace}
\usepackage{algorithm}
\usepackage{algorithmic}

\definecolor{sensorblue}{RGB}{66,133,244}
\definecolor{alertred}{RGB}{219,68,55}
\definecolor{contextgreen}{RGB}{15,157,88}
\definecolor{reportgold}{RGB}{244,180,0}
\definecolor{lightgray}{RGB}{245,245,245}

\title{\textbf{Hierarchical Attention-based Graph Neural Network\\with Relevance-driven Pruning}}

\author{
Seungwoo Kum\\
Korea Electronics Technology Institute (KETI)\\
Seongnam, Republic of Korea\\
\texttt{swkum@keti.re.kr}
}

\date{}

\begin{document}

\maketitle

\begin{abstract}
Graph Neural Networks (GNNs) excel at relational reasoning but face two persistent challenges: the lack of interpretable attribution for heterogeneous node types, and the computational overhead of message passing over large, noisy graphs. We propose the Hierarchical Attention-based Heterogeneous GNN (HA-HeteroGNN), a framework that addresses both issues through a unified explainability-to-pruning pipeline. A two-tier attention mechanism separates sensor-level and context-level computation across 16 node types and 18 edge types, producing per-node relevance scores via an attention-based GNN Explainer without requiring gradient backpropagation. These relevance scores then serve as a principled pruning criterion: removing nodes identified as consistently uninformative yields a 27\% reduction in graph edges while simultaneously improving classification accuracy by 2.4--6.1\% across all model variants, challenging the conventional assumption that pruning necessarily trades accuracy for efficiency. Experiments on a 50,000-record synthetic dataset spanning 11 report categories demonstrate 97.5\% cross-strategy explanation stability and domain consistent sensor attribution, with training-time reductions of up to 43.9\% and real-time inference latency of approximately 58--60\,ms per sample.

\end{abstract}

\section{Introduction}
\label{sec:intro}

Governments worldwide operate citizen reporting platforms to handle non-emergency public safety concerns. South Korea's \textit{Safety e-Report} is a representative system that enables citizens to report safety hazards including infrastructure damage, environmental concerns, and weather related risks through mobile and web interfaces~\citep{korea_safety_2020}. Similarly, North America's 311 service provides a unified channel for non-emergency municipal requests across hundreds of cities in the United States and Canada~\citep{nam2014311, minkoff2016nyc311}. The United Kingdom's FixMyStreet platform~\citep{fixmystreet2023} and Australia's Snap Send Solve~\citep{snapsendsolve2022} serve analogous functions in their respective jurisdictions. These systems share a common characteristic: reports are classified as non-emergency at the point of intake, under the assumption that they do not require immediate response. However, this static classification fails to account for rapidly evolving environmental conditions that can transform routine reports into urgent situations.

Consider the following scenarios where non-emergency reports acquire unexpected urgency: a drainage maintenance request filed during a period when 80\,mm of rainfall is recorded within one hour under a heavy rain alert; a leaf cleanup request coinciding with a sudden downpour that creates flash flooding conditions in the reported area; a road icing prevention request filed just as temperatures plummet below $-15^\circ$C with a cold wave alert approaching; or a fine dust complaint submitted during a severe yellow dust event with PM$_{2.5}$ concentrations exceeding 300\,$\mu$g/m$^3$. In each case, the urgency cannot be determined from the report data alone--it requires integration of external information including real-time sensor readings, meteorological alerts, and spatial context. This \textit{urgency gap} between static classification and dynamic situational awareness represents a critical challenge for public safety operations.

The urgency assessment problem is inherently relational: the risk level of a report depends on the relationships between the report's category, its geographic location, current weather conditions, active meteorological alerts, and temporal patterns. While various deep learning models could be applied to this problem, the relational structure naturally maps to a graph representation where entities become nodes and their interactions become edges. Graph Neural Networks (GNNs) have demonstrated remarkable success in modeling such relational data. \citet{kipf2017semi} introduced Graph Convolutional Networks (GCN) for semi-supervised classification on graph-structured data. \citet{hamilton2017inductive} proposed GraphSAGE, enabling inductive learning on large scale graphs through neighborhood sampling and aggregation. \citet{velickovic2018graph} developed Graph Attention Networks (GAT) with attention mechanisms for adaptive neighbor weighting. For heterogeneous graphs containing multiple node and edge types, \citet{wang2019heterogeneous} proposed Heterogeneous Graph Attention Networks (HAN) and \citet{hu2020heterogeneous} introduced Heterogeneous Graph Transformer (HGT). In the domain of emergency management, GNNs have been applied to traffic incident prediction~\citep{yu2018spatio}, urban event detection~\citep{zhang2020urban_event}, and disaster response optimization~\citep{fan2019disaster_gnn}, demonstrating the suitability of graph-based approaches for capturing complex interdependencies.

Despite their effectiveness, GNNs face significant scalability challenges. As graph sizes increase with growing data volumes, both training and inference demand substantial computational resources. To address training scalability, \citet{hamilton2017inductive} proposed the SAGE (Sample and AggregatE) framework which samples a fixed-size neighborhood rather than using the entire graph. \citet{chen2018fastgcn} introduced FastGCN with importance sampling, \citet{chiang2019clustergcn} proposed Cluster-GCN for efficient minibatch training, and \citet{zeng2020graphsaint} developed GraphSAINT for subgraph-based sampling. A further critical distinction arises between \textit{transductive} and \textit{inductive} inference paradigms. Transductive models~\citep{kipf2017semi} operate on a fixed graph and cannot generalize to unseen nodes without retraining on the entire graph. Inductive models~\citep{hamilton2017inductive}, in contrast, can generate embeddings for previously unseen nodes by aggregating information from their local neighborhoods. For continuously arriving incident reports, inductive inference is essential--each new report must be classified without rebuilding the entire graph. However, inductive inference introduces its own challenge: when a single new report is converted into a graph, the resulting subgraph is extremely small (typically one report node with a handful of sensor and context nodes), providing insufficient structural context for accurate classification.

To address these challenges, we propose the \textbf{Hierarchical Attention-based Heterogeneous GNN (HA-HeteroGNN)} framework with the following five contributions:

\begin{enumerate}[leftmargin=2em, itemsep=1pt]
\item \textbf{Constraint-based Synthetic Data Generation}: An algorithm for generating realistic multimodal emergency scenarios with enforced spatiotemporal consistency, alert-weather constraints, and seasonal patterns (Section~\ref{sec:datagen}).
\item \textbf{Anchor-based Input Graph Construction}: A technique for constructing representative anchor graphs that overcome the cold start problem in inductive GNN inference (Section~\ref{sec:anchor}).
\item \textbf{Hierarchical Multihead Attention}: A two-tier attention architecture that separates sensor-level and context-level attention computation, enabling fine grained attribution across heterogeneous data modalities (Section~\ref{sec:attention}).
\item \textbf{Attention-based GNN Explainer}: An explainability mechanism that leverages learned attention weights to quantify the influence of each node type on urgency predictions, without requiring computationally expensive gradient backpropagation (Section~\ref{sec:explainer}).
\item \textbf{Relevance-driven Graph Pruning}: A graph pruning strategy guided by explainability scores that removes low influence edges and nodes, simultaneously improving accuracy and reducing computational cost (Section~\ref{sec:pruning}).
\end{enumerate}

The remainder of this paper is organized as follows. Section~\ref{sec:related} reviews related work in GNN architectures, explainability, and graph optimization. Section~\ref{sec:proposed} presents the proposed HA-HeteroGNN framework as a five-stage pipeline from data generation through graph pruning. Section~\ref{sec:implementation} describes the implementation details, including graph structure and model architectures. Section~\ref{sec:experiments} presents experimental results and analysis. Section~\ref{sec:conclusion} concludes the paper and discusses future directions.

\section{Related Work}
\label{sec:related}

Graph Neural Networks have evolved rapidly since their introduction by \citet{scarselli2009graph}. The spectral approach by \citet{bruna2014spectral} laid the theoretical foundation, followed by the computationally efficient ChebNet~\citep{defferrard2016convolutional} and GCN~\citep{kipf2017semi}. The spatial approach, exemplified by GraphSAGE~\citep{hamilton2017inductive} and Message Passing Neural Networks (MPNN)~\citep{gilmer2017neural}, operates directly on graph topology through neighborhood aggregation. For heterogeneous graphs, Relational Graph Convolutional Networks (R-GCN)~\citep{schlichtkrull2018modeling} apply type specific weight matrices, Heterogeneous Graph Attention Networks (HAN)~\citep{wang2019heterogeneous} use metapath-based attention for semantic-level aggregation, and Heterogeneous Graph Transformer (HGT)~\citep{hu2020heterogeneous} uses type dependent parameters for attention computation. More recently, Simple-HGN~\citep{lv2021simple_hgn} demonstrated that careful architecture design can achieve competitive performance without metapaths. In the context of urban computing and public safety, \citet{geng2019spatiotemporal} applied multigraph convolution for ride-hailing demand prediction and \citet{zheng2020gman} proposed GMAN for traffic flow prediction using spatial-temporal attention.

Explainability in GNNs has become critical for safety-critical applications. GNNExplainer~\citep{ying2019gnnexplainer} identifies important subgraph structures and node features through mutual information maximization. PGExplainer~\citep{luo2020parameterized} learns a parameterized mask generator for edge importance. SubgraphX~\citep{yuan2021subgraphx} employs Monte Carlo tree search to identify explanatory subgraphs. Gradient-based methods adapted from computer vision, including Integrated Gradients~\citep{sundararajan2017axiomatic} and Grad-CAM for GNNs~\citep{pope2019explainability_gnn}, provide feature-level importance scores through backpropagation. However, these post-hoc methods incur significant computational overhead and may produce inconsistent explanations across similar inputs. Attention-based explainability offers an alternative where importance scores are intrinsic to the model's computation--\citet{velickovic2018graph} showed that GAT attention weights can serve as proxies for edge importance, and \citet{brody2022how} analyzed the expressiveness of different attention formulations in GNNs.

The scalability of GNNs to large graphs has motivated extensive research on memory optimization. Sampling-based methods including GraphSAGE~\citep{hamilton2017inductive}, FastGCN~\citep{chen2018fastgcn}, and GraphSAINT~\citep{zeng2020graphsaint} reduce memory consumption by operating on subgraphs. Graph sparsification methods directly reduce graph size: \citet{zheng2020robust_graph_sparsification} proposed spectral sparsification, \citet{chen2021unified_lottery} applied the lottery ticket hypothesis to identify sparse GNN subnetworks, and DropEdge~\citep{rong2019dropedge} randomly removes edges during training as regularization. Knowledge distillation~\citep{yang2020distilling_gnn} compresses large GNN models, while graph coarsening~\citep{huang2021scaling_graph_coarsening} reduces graph size preserving structural properties. These methods focus on general optimization without considering the semantic relevance of individual nodes.

For inductive inference, GraphSAGE~\citep{hamilton2017inductive} enabled learning on new nodes by sampling and aggregating local neighborhoods, and PinSage~\citep{ying2018pinsage} demonstrated scalability to billion node graphs. For new isolated nodes, the quality of the input graph significantly affects prediction accuracy. \citet{fatemi2021slaps} proposed learning graph structure jointly with node representations. The challenge is particularly acute when individual input instances produce very small graphs with limited structural information. To our knowledge, no prior work has specifically addressed the construction of anchor graphs for enriching single-node inductive inference in heterogeneous GNN settings.

\section{Proposed Method: HA-HeteroGNN Framework}
\label{sec:proposed}

\begin{figure}[t]
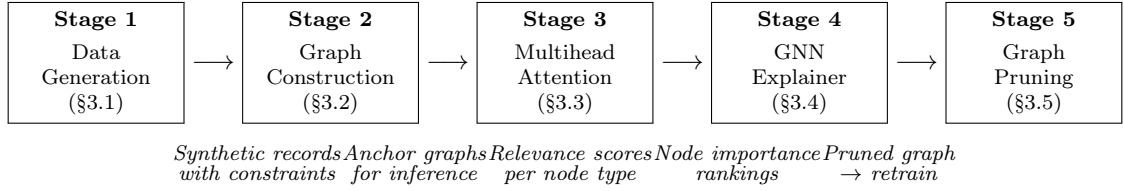

\centering
\small
\setlength{\tabcolsep}{0pt}
\begin{tabular}{ccccccccc}
\fbox{\parbox{2.1cm}{\centering\scriptsize\textbf{Stage 1}\\\smallskip Data\\Generation\\(\S\ref{sec:datagen})}} &
$\;\longrightarrow\;$ &
\fbox{\parbox{2.1cm}{\centering\scriptsize\textbf{Stage 2}\\\smallskip Graph\\Construction\\(\S\ref{sec:anchor})}} &
$\;\longrightarrow\;$ &
\fbox{\parbox{2.1cm}{\centering\scriptsize\textbf{Stage 3}\\\smallskip Multihead\\Attention\\(\S\ref{sec:attention})}} &
$\;\longrightarrow\;$ &
\fbox{\parbox{2.1cm}{\centering\scriptsize\textbf{Stage 4}\\\smallskip GNN\\Explainer\\(\S\ref{sec:explainer})}} &
$\;\longrightarrow\;$ &
\fbox{\parbox{2.1cm}{\centering\scriptsize\textbf{Stage 5}\\\smallskip Graph\\Pruning\\(\S\ref{sec:pruning})}}
\end{tabular}

\medskip
\scriptsize
\begin{tabular}{ccccc}
\textit{Synthetic records} & \textit{Anchor graphs} & \textit{Relevance scores} & \textit{Node importance} & \textit{Pruned graph} \\[-2pt]
\textit{with constraints} & \textit{for inference} & \textit{per node type} & \textit{rankings} & \textit{$\to$ retrain}
\end{tabular}
\caption{HA-HeteroGNN pipeline. Each stage feeds its output to the next; the pruned graph from Stage~5 is used to retrain the model from scratch, closing the explainability-to-pruning loop.}
\label{fig:pipeline}
\end{figure}

\begin{figure}[t]
\centering
\includegraphics[width=0.75\textwidth]{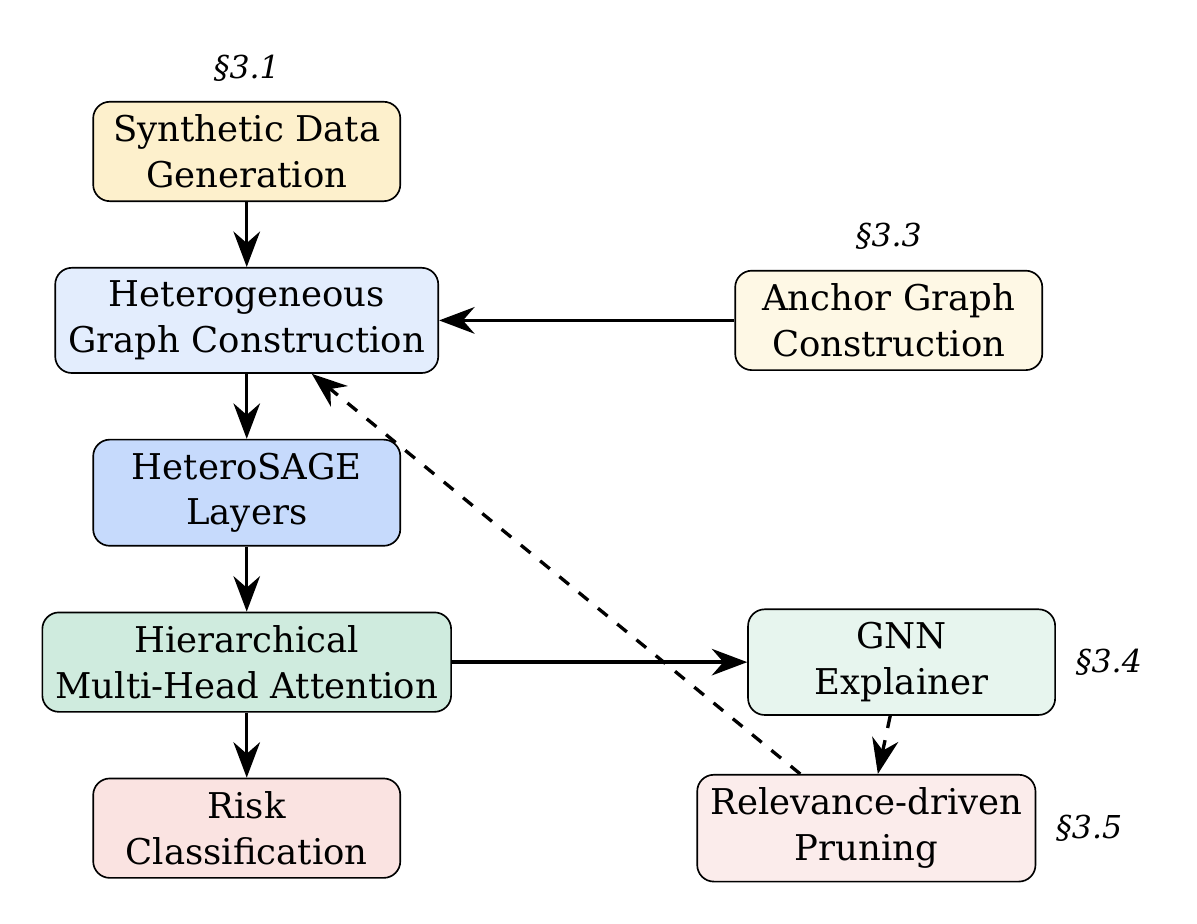}
\caption{Overview of the HA-HeteroGNN framework. Solid arrows indicate the primary data flow; dashed arrows indicate the feedback loop from explainability driven pruning.}
\label{fig:framework_overview}
\end{figure}

We present the \textbf{Hierarchical Attention-based Heterogeneous Graph Neural Network (HA-HeteroGNN)} framework, whose central objective is to improve both accuracy and efficiency of graph based inference through \textit{relevance driven graph pruning}. The key insight is that hierarchical attention over heterogeneous node types produces per-node relevance scores that can serve as a principled pruning criterion, enabling the removal of uninformative edges from the input graph before retraining. The framework operates as a five-stage pipeline (Figure~\ref{fig:pipeline}): (1)~constraint-based synthetic data generation, (2)~anchor-based input graph construction for inductive inference, (3)~hierarchical multihead attention for modality-separated message passing, (4)~attention-based GNN Explainer for node relevance quantification, and (5)~relevance driven graph pruning. Figure~\ref{fig:framework_overview} shows how these stages are integrated in the system architecture.

\subsection{Constraint-based Synthetic Data Generation}
\label{sec:datagen}

The urgency of a non-emergency report depends on the interplay between the report category, environmental sensor readings, meteorological alert status, and spatial context. We model these relationships as a heterogeneous graph $\mathcal{G} = (\mathcal{V}, \mathcal{E})$ containing multiple node types: sensor nodes representing environmental measurements (rainfall, temperature, humidity, wind speed, snowfall, PM$_{2.5}$, apparent temperature), alert-related nodes capturing meteorological warning status, and contextual nodes encoding spatial and structural information. Each report node is connected to its associated sensor, alert, and context nodes via typed edges, forming a rich relational structure. The specific node type encodings and graph topology are detailed in Section~\ref{sec:implementation}.

Generating realistic synthetic data requires enforcing three types of consistency constraints. First, \textbf{alert-weather constraints} link 12 alert types to physical sensor value ranges: for example, a heavy rain warning requires rainfall of 90--150\,mm, a cold wave warning requires temperatures between $-25^\circ$C and $-15^\circ$C, and a yellow dust warning requires PM$_{2.5}$ of 400--800\,$\mu$g/m$^3$. Second, \textbf{alert sequence consistency} enforces three progression scenarios: direct alert issuance, pre-alert followed by alert, and pre-alert followed by escalation (advisory to warning). Third, \textbf{spatiotemporal coherence} constrains sensor readings within a one-hour time window across adjacent districts to vary by at most $\pm15\%$, implemented through a \texttt{TemporalSensorManager} that maintains a 24-hour rolling window of readings per location.

The complete generation algorithm is formalized in Algorithm~\ref{alg:datagen}.

\begin{algorithm}[t]
\caption{Constraint-based Synthetic Data Generation}
\label{alg:datagen}
\small
\begin{algorithmic}[1]
\REQUIRE $N$ (number of records), year, risk\_distribution
\ENSURE Dataset $D = \{(r_i, s_i, a_i, l_i, y_i)\}$
\STATE Initialize TemporalSensorManager (24-hour rolling window)
\STATE Initialize AlertManager (alert sequence tracker)
\FOR{$i = 1$ to $N$}
  \STATE report\_type $\leftarrow$ sample from REPORT\_TYPES
  \STATE season $\leftarrow$ get\_season(report\_type)
  \STATE timestamp $\leftarrow$ generate\_timestamp(year, season)
  \STATE risk\_level $\leftarrow$ sample from risk\_distribution
  \IF{risk\_level $\in$ \{medium, high\}}
    \STATE (alert, pre\_alert) $\leftarrow$ generate\_alert\_sequence(report\_type, risk\_level)
  \ELSE
    \STATE (alert, pre\_alert) $\leftarrow$ (none, none)
  \ENDIF
  \IF{alert $\neq$ none}
    \STATE sensors $\leftarrow$ generate\_constrained\_sensors(alert)
  \ELSE
    \STATE sensors $\leftarrow$ generate\_risk\_based\_sensors(report\_type)
  \ENDIF
  \STATE location $\leftarrow$ sample\_location(season)
  \STATE sensors $\leftarrow$ enforce\_temporal\_consistency(sensors, location, timestamp, $\sigma{=}0.15$)
  \STATE $D[i] \leftarrow$ (report\_type, sensors, alert, location, risk)
\ENDFOR
\RETURN $D$
\end{algorithmic}
\end{algorithm}

\subsection{Anchor-based Input Graph Construction}
\label{sec:anchor}

When processing non-emergency reports with a GNN in a production setting, each incoming report must be converted into a graph for inference. In a transductive setting, the new report would be added to the existing graph and the entire model retrained--computationally prohibitive for real-time applications. Inductive models such as our HeteroSAGE backbone can process new nodes without retraining. However, when a single report is converted to a graph in isolation, the resulting structure contains typically one report node connected to 7--12 sensor/context nodes. This minimal graph lacks the structural diversity and inter-node relationships that the model learned during training on graphs containing thousands of nodes, leading to degraded inference quality.

We propose three anchor strategies that augment the input graph with representative training data. Table~\ref{tab:anchor_strategies} summarizes these strategies along with the no-anchor baseline.

\begin{table}[t]
\centering
\caption{Anchor graph construction strategies for inductive inference.}
\label{tab:anchor_strategies}
\small
\begin{tabularx}{\textwidth}{lXll}
\toprule
\textbf{Strategy} & \textbf{Description} & \textbf{Latency} & \textbf{Context} \\
\midrule
Single-Node & No anchor; input report embedded alone (baseline). & $\sim$58\,ms & Minimal \\
Synthetic & Synthetic anchor generated from learned prototypes via generative sampling. & $\sim$239\,ms & Rich \\
Median & Per-group median training record as anchor, providing a statistically central reference. & $\sim$60\,ms & Moderate \\
Coverage & $k{=}5$ representative anchors per group, selected via farthest point sampling to maximize feature space coverage. & $\sim$60\,ms & Diverse \\
\bottomrule
\end{tabularx}
\end{table}

Anchor nodes are pre-computed during training and stored alongside the model checkpoint. At inference time, anchors are selected based on the incoming report's type and combined with the new report to form the input graph.

\subsection{Hierarchical Multihead Attention}
\label{sec:attention}

The urgency assessment graph contains fundamentally different types of information: continuous sensor measurements (rainfall, temperature, humidity), categorical alert statuses (weather warnings, pre-alerts), and structural context (location, drainage condition). When a flat attention mechanism is applied uniformly across all node types, these distinct modalities are conflated, making it difficult to disentangle the influence of sensor readings from contextual factors. For example, when a heavy snow incident report is classified as high risk, a domain expert would naturally ask: ``Is this primarily due to the snowfall amount, or because a snow alert has been issued?'' A single attention computation mixing all node types cannot cleanly separate these influences.

We propose a \textbf{hierarchical multihead attention} mechanism operating at two levels. At \textbf{Level 1 (Intra-Category Attention)}, three specialized attention modules operate on distinct subsets of neighbor nodes: (i) the \textbf{Sensor Head} $\mathcal{A}_S$ attends only to the 7 sensor node types (rainfall, temperature, humidity, apparent temperature, wind speed, PM, snowfall); (ii) the \textbf{Context Head} $\mathcal{A}_C$ attends to contextual nodes (weather alert, pre-alert type/time/severity, location, drainage condition); and (iii) the \textbf{Full Head} $\mathcal{A}_F$ attends to all neighbor node types including report count and report type. Each head internally uses 4-head multihead attention with $d=128$ dimensions. At \textbf{Level 2 (Intercategory Fusion)}, the outputs of the three heads are concatenated and projected through a learned combiner: $[128\times3] \to [256] \to \text{ReLU} \to \text{LayerNorm} \to [128]$.

The attention query is conditioned on the report type to enable category specific patterns. Let $\mathbf{h}_r \in \mathbb{R}^d$ be the report embedding after GNN message passing and $\mathbf{h}_{rt}$ be its report type embedding. The conditioned query is $\mathbf{q}_r = \mathbf{h}_r + \mathbf{h}_{rt}$, allowing the model to learn that a heavy snow report should attend more to snowfall sensors while a fine dust report attends more to PM sensors. For each head $k \in \{S, C, F\}$ with allowed node types $\mathcal{T}_k$:
\begin{equation}
\alpha_{r,j}^{(k)} = \frac{\exp\!\bigl(\mathbf{q}_r^\top \mathbf{W}_Q^{(k)} (\mathbf{W}_K^{(k)} \mathbf{h}_{v_j})^\top / \sqrt{d/H}\bigr)}{\sum_{j': t_{j'} \in \mathcal{T}_k} \exp\!\bigl(\mathbf{q}_r^\top \mathbf{W}_Q^{(k)} (\mathbf{W}_K^{(k)} \mathbf{h}_{v_{j'}})^\top / \sqrt{d/H}\bigr)}, \quad
\mathbf{o}_r^{(k)} = \text{LN}\!\left(\mathbf{h}_r + \textstyle\sum_{j} \alpha_{r,j}^{(k)} \mathbf{W}_V^{(k)} \mathbf{h}_{v_j}\right)
\end{equation}
where LN denotes LayerNorm and $H{=}4$ is the number of internal heads. The final risk prediction is $\hat{y}_r = \text{softmax}\!\bigl(\text{MLP}_{\text{cls}}(\text{MLP}_{\text{fuse}}([\mathbf{o}_r^{(S)} \| \mathbf{o}_r^{(C)} \| \mathbf{o}_r^{(F)}]))\bigr)$.

\begin{figure}[t]
\centering
\includegraphics[width=0.65\textwidth]{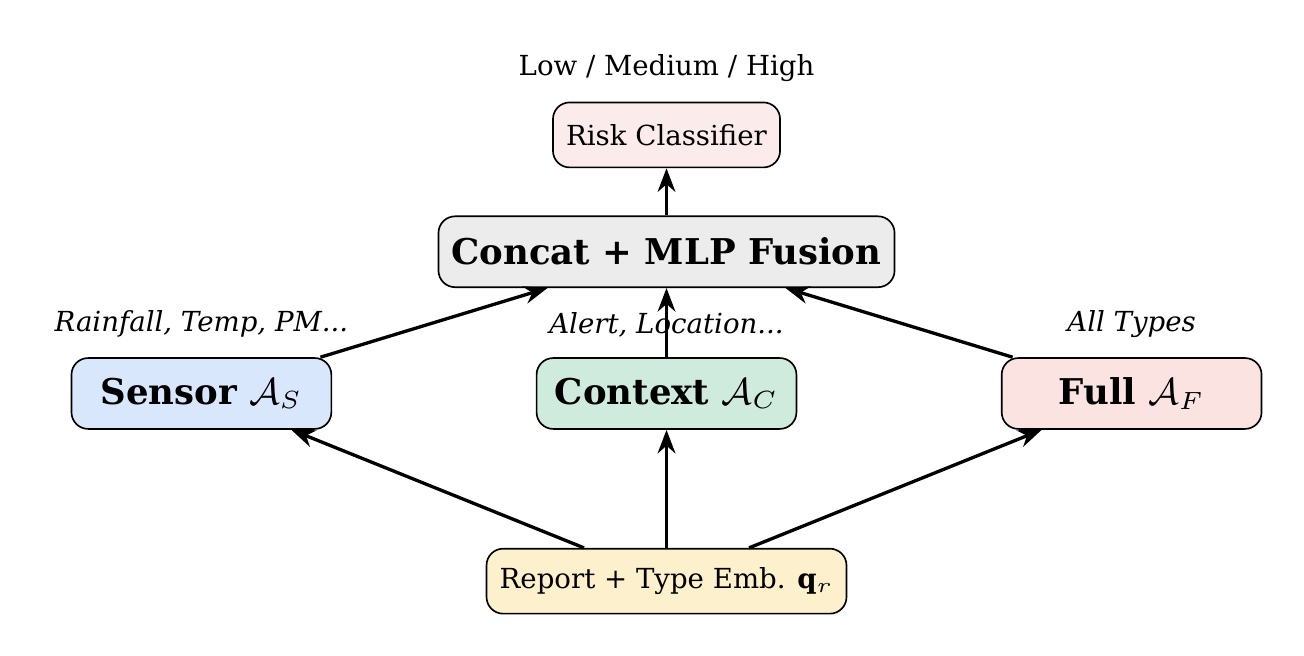}
\caption{Hierarchical multihead attention architecture. The report type conditioned query $\mathbf{q}_r$ is fed to three specialized attention heads, whose outputs are fused through concatenation and MLP projection.}
\label{fig:hierarchical_attention}
\end{figure}

Figure~\ref{fig:hierarchical_attention} illustrates the architecture.

\subsection{Attention-based GNN Explainer}
\label{sec:explainer}

Because GNN models encode relational structure, the influence of each node on the final prediction can be quantified. Two primary approaches exist. \textbf{Gradient-based methods} compute feature-level importance through backpropagation: $I_{\text{grad}}(v_j) = |\partial \hat{y}_r / \partial \mathbf{h}_{v_j}| \cdot |\mathbf{h}_{v_j}|$. While theoretically grounded, gradient methods require a backward pass for each explanation and may suffer from gradient saturation. \textbf{Attention-based methods} directly use learned attention weights: $I_{\text{attn}}(v_j) = \sum_{k} w_k \cdot \alpha_{r,j}^{(k)}$, requiring no additional computation beyond the forward pass.

While attention-based approaches generally achieve higher accuracy in node importance estimation, our framework specifically requires \textit{independent per-report-type} node attributions--the same sensor readings should receive different importance depending on whether the report concerns heavy snow versus fine dust. This motivates our combination of HeteroSAGE message passing (which provides strong structural embeddings) with report type conditioned attention (which provides category specific explanations). The attention weights from the forward pass serve directly as explainability scores, with \texttt{location}, \texttt{report\_count}, and \texttt{report\_type} excluded from importance ranking to avoid trivially dominant nodes. The mean importance across a sample set $\mathcal{S}$ is: $\bar{I}(\text{type}) = (1/|\mathcal{S}|) \sum_{s \in \mathcal{S}} I_s(\text{type})$, normalized to sum to 100\%.

\begin{figure}[t]
\centering
\includegraphics[width=0.8\textwidth]{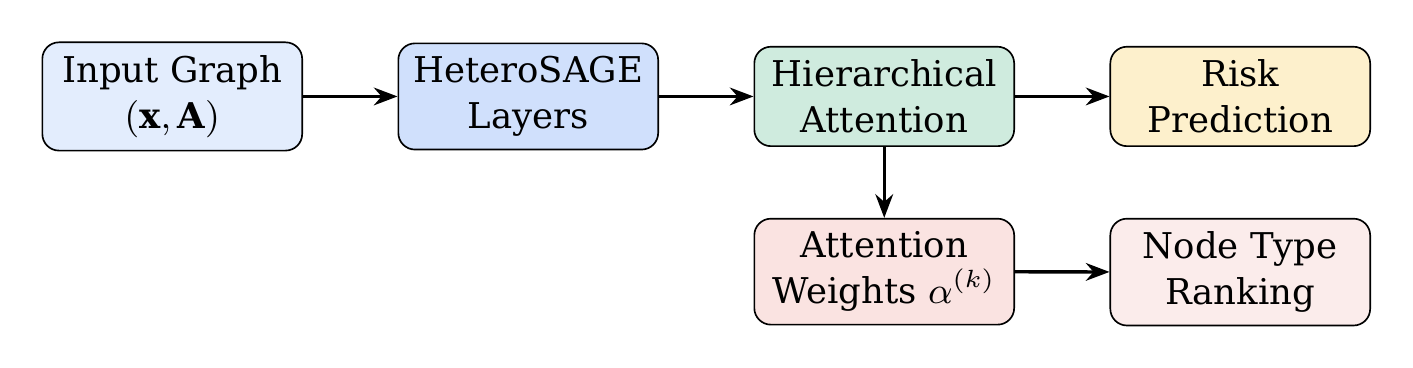}
\caption{GNN Explainer architecture. Attention weights from the hierarchical attention module produce per-node-type importance rankings without additional backward passes.}
\label{fig:explainer}
\end{figure}

Figure~\ref{fig:explainer} illustrates the GNN Explainer pipeline.

\subsection{Relevance-driven Graph Pruning}
\label{sec:pruning}

The GNN Explainer (Section~\ref{sec:explainer}) produces per-node-type importance scores that quantify each node's contribution to the final risk prediction. We exploit these scores to construct pruned graph variants by selectively removing edges associated with low importance nodes. The procedure consists of three steps: (1)~run the Explainer on a representative sample from the training set to compute mean importance $\bar{I}(\text{type})$ per node type, stratified by report category and risk level; (2)~identify nodes that are consistently low importance across all risk levels within each report category (i.e., nodes whose $\bar{I}$ falls outside the top-3 ranking for every risk level); and (3)~remove all edges connecting the report node to these low importance node types.

We define two pruning strategies of increasing aggressiveness:

\begin{itemize}[leftmargin=2em, itemsep=1pt]
\item \textbf{Bottom-excluded}: Remove edges to node types that rank consistently low across \textit{all} risk levels for each report category. This preserves nodes that are important in any risk scenario, yielding a conservative pruning that retains domain relevant connections.
\item \textbf{Top-only}: Retain edges \textit{only} to the top-3 most important node types per (report category, risk level) combination. This aggressive strategy maximizes graph compression but risks discarding nodes that carry secondary but non-negligible predictive signals.
\end{itemize}

This approach differs from existing graph sparsification methods~\citep{zheng2020robust_graph_sparsification, rong2019dropedge} in two respects. First, the pruning criterion derives from the model's own learned attention patterns rather than from random sampling or spectral properties. Second, pruning is \textit{category-conditioned}--the same node type may be retained for one report category (e.g., Rainfall for flood incidents) and removed for another (e.g., Rainfall for fine dust reports), enabling semantically targeted graph compression. Models are retrained from scratch on the pruned graphs to fully adapt to the modified topology.

\section{Implementation}
\label{sec:implementation}

\subsection{Dataset Generation}

We generated a synthetic dataset of \textbf{50,040 records} spanning 11 emergency report categories across four seasonal groups: Water/Flood (Leaf Cleanup, Drainage Maintenance, Flood Prevention, Landslide Risk), Cold/Snow (Road Icing Prevention, Heavy Snow, Cold Wave), Heat (Heat Wave), and Air Quality (Fine Dust Report, Wildfire Prevention, Yellow Dust). The risk level distribution follows 25\%/35\%/40\% for Low/Medium/High, yielding 1,321--1,855 samples per (report type, risk level) group. Each record contains 17 fields: report timestamp, classification, location (among 9 Seoul districts across 3 regions), 7 sensor values, weather alert type, pre-alert type/time/severity, drainage status, and hourly co-located report count. Data achieves 100\% completeness with 93.7\% constraint satisfaction across 12 alert-weather rules. The dataset is split into 80\%/10\%/10\% for training, validation, and test sets using stratified sampling to preserve the (report type, risk level) distribution. Within the training set, we use a \textit{labeled ratio} of 20\%, meaning only 20\% of training nodes have observed labels during GNN message passing, simulating a semi-supervised setting where labeled incident reports are scarce relative to the volume of incoming data.

\subsection{Graph Structure}

\begin{figure}[t]
\centering
\includegraphics[width=0.7\textwidth]{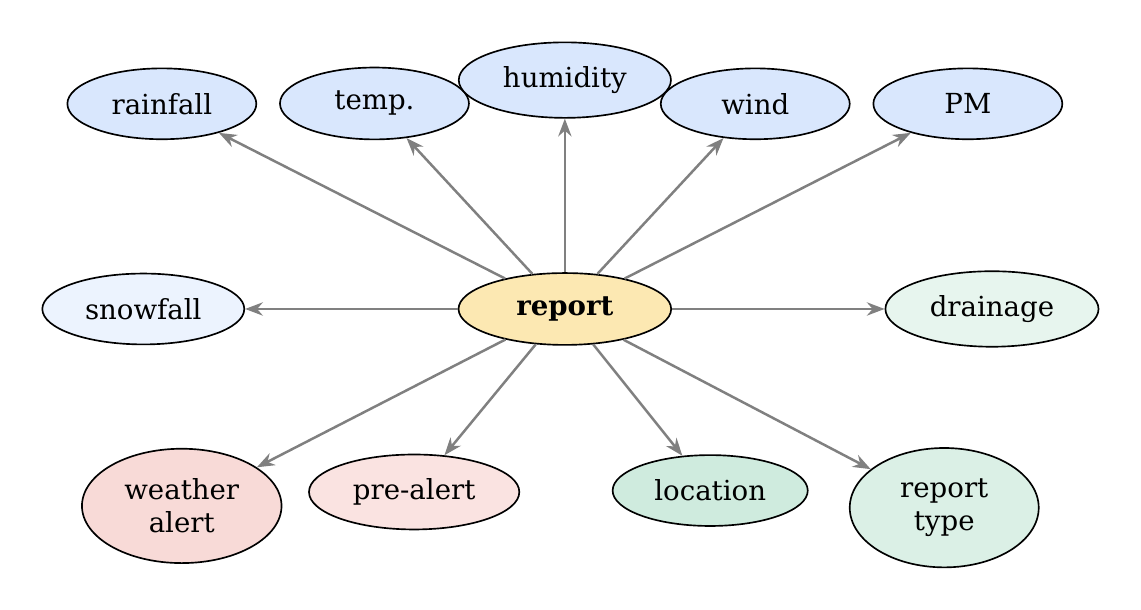}
\caption{Heterogeneous graph schema. Each report node connects to sensor, alert, and context nodes. Location nodes additionally form spatial adjacency edges (not shown).}
\label{fig:graph_schema}
\end{figure}

The heterogeneous graph consists of 16 node types and 18 bidirectional edge types. Each report node connects to 7 sensor nodes (rainfall, temperature, apparent temperature, humidity, wind speed, snowfall, PM$_{2.5}$), 4 alert-related nodes (weather alert type, pre-alert type/time/severity), and contextual nodes (location, drainage condition, co-located report count, report category). Sensor nodes use a 6-dimensional encoding combining 5-bin categorical quantization with a normalized continuous value. Alert nodes use one-hot encodings of 13 alert types. Location nodes encode latitude, longitude, region code, and distance from center in 4 dimensions. Figure~\ref{fig:graph_schema} presents the complete schema.

\subsection{Model Architectures}

\begin{figure}[t]
\centering
\includegraphics[width=0.85\textwidth]{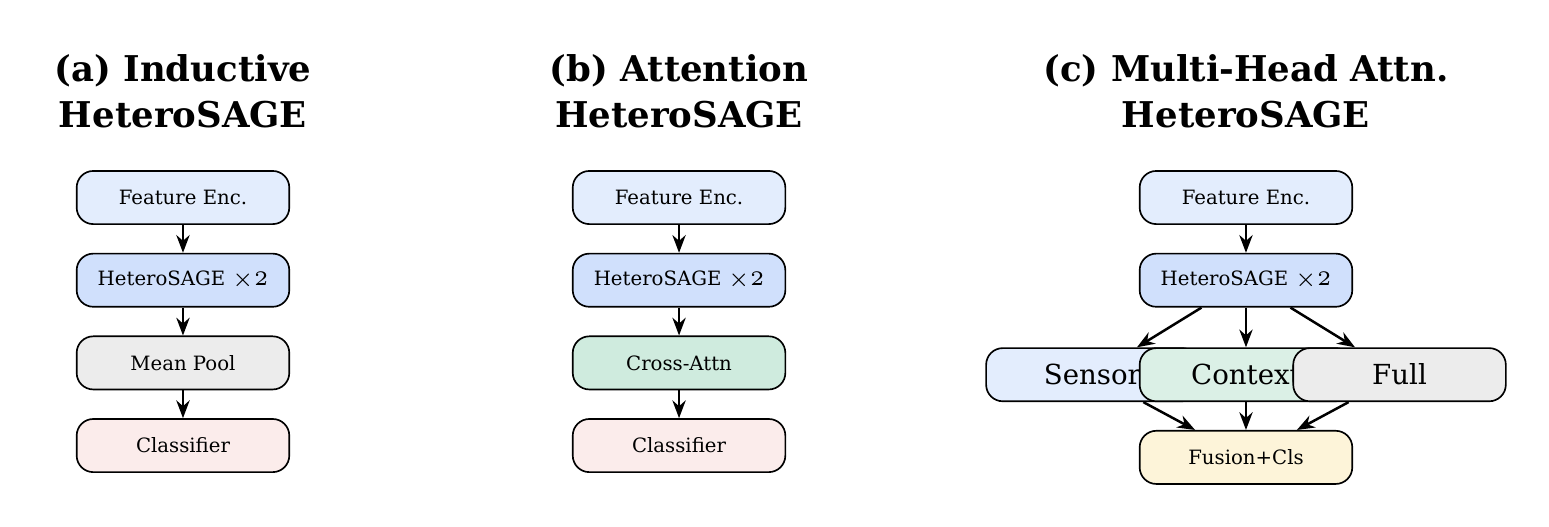}
\caption{Three model architectures: (a) base inductive without attention; (b) single cross-attention; (c) proposed hierarchical multihead attention.}
\label{fig:model_comparison}
\end{figure}

We implement and compare three model variants sharing the same backbone: 2-layer HeteroSAGE with $d{=}128$ hidden dimensions, mean aggregation, LayerNorm, residual connections, and dropout ($p{=}0.3$). Training uses CrossEntropyLoss with inverse-frequency class weights, Adam optimizer (lr${}=10^{-3}$), and early stopping (patience=20). Table~\ref{tab:models} summarizes the architectures, and Figure~\ref{fig:model_comparison} illustrates their structural differences.

\begin{table}[t]
\centering
\caption{Model architectures compared in this study.}
\label{tab:models}
\small
\begin{tabularx}{\textwidth}{lXcc}
\toprule
\textbf{Model} & \textbf{Description} & \textbf{Params} & \textbf{Explain.} \\
\midrule
Inductive HeteroSAGE & Base model with SAGEConv + mean aggregation. No attention. & $\sim$180K & Gradient \\
Attention HeteroSAGE & Single cross-attention head over all neighbor types with report type conditioned query. & $\sim$350K & Attention \\
Multihead Attn. HeteroSAGE & \textbf{Proposed.} Three specialized heads (Sensor/Context/Full) with intercategory fusion. & $\sim$520K & Hierarchical \\
\bottomrule
\end{tabularx}
\end{table}

\subsection{Evaluation Metrics}

Our primary evaluation focus is \textit{explainability quality} rather than raw classification accuracy, as the goal is to identify which environmental factors contribute most to urgency elevation. We evaluate: (i) \textbf{node-type importance scores}--mean attention-weight-based importance (\%) per node type, normalized to 100\% per sample; (ii) \textbf{cross-strategy agreement}--the fraction of samples for which all four anchor strategies identify the same top-1 most important node type; (iii) \textbf{domain appropriateness}--whether identified important nodes align with physical domain knowledge; (iv) \textbf{inference latency}--per-sample wall-clock time including graph construction and inference. For high risk reports, we aim to identify high-influence nodes (candidates for retention) and low influence nodes (candidates for graph pruning).

\subsection{Experimental Design}

We evaluate the interaction between three model architectures and four anchor strategies (Table~\ref{tab:anchor_strategies}). For each combination, we perform stratified sampling of 100 records per (report type, risk level) group, yielding $11 \times 3 \times 100 = 3{,}300$ samples per strategy. The primary experiment uses the Attention HeteroSAGE model (trained on the full dataset for 10 epochs with 20\% labeled ratio) with the Coverage anchor strategy to establish baseline explainability results, followed by comparison across strategies.

\section{Experimental Results}
\label{sec:experiments}

\subsection{Inference Latency}

Table~\ref{tab:inference_time} reports per-sample inference latency measured over 3,300 samples on a single GPU (\texttt{cuda:2}). Single-Node, Median, and Coverage strategies achieve comparable latency ($\approx$58--60\,ms), while Synthetic is approximately 4$\times$ slower due to real-time prototype generation. This demonstrates that pre-computed anchor strategies provide enriched context with negligible overhead.

\begin{table}[t]
\centering
\caption{Per-sample inference latency (ms) over 3,300 samples.}
\label{tab:inference_time}
\small
\begin{tabular}{lccc}
\toprule
\textbf{Strategy} & \textbf{Mean (ms)} & \textbf{Std (ms)} & \textbf{Median (ms)} \\
\midrule
Single-Node & 57.6 & 18.0 & 53.8 \\
Synthetic & 239.3 & 40.6 & 228.4 \\
Median & 59.6 & 17.8 & 56.3 \\
Coverage & 60.0 & 18.4 & 56.1 \\
\bottomrule
\end{tabular}
\end{table}

\subsection{Cross-Strategy Robustness}

We measure the \textbf{top-1 node agreement rate}: the fraction of samples for which all four anchor strategies produce the same most-important node type. Table~\ref{tab:agreement} presents results by risk level and by report type.

\begin{table}[t]
\centering
\caption{Top-1 node type agreement rate (\%) across four anchor strategies.}
\label{tab:agreement}
\small
\begin{tabular}{lclc}
\toprule
\textbf{By Risk Level} & \textbf{Agree (\%)} & \textbf{By Report Type} & \textbf{Agree (\%)} \\
\midrule
Low Risk & 95.1 & Leaf Cleanup & 98.7 \\
Medium Risk & 99.4 & Road Icing Prev. & 99.7 \\
High Risk & 98.1 & Fine Dust Report & 99.7 \\
\cmidrule{1-2}
\textbf{Overall} & \textbf{97.5} & Drainage Maint. & 99.0 \\
 & & Wildfire Prev. & 94.7 \\
 & & Landslide Risk & 92.0 \\
 & & Flood Prev. & 95.3 \\
 & & Heavy Snow & 100.0 \\
 & & Heat Wave & 99.3 \\
 & & Cold Wave & 95.7 \\
 & & Yellow Dust & 98.7 \\
\bottomrule
\end{tabular}
\end{table}

The overall agreement of 97.5\% ($\pm$1.2\% across three independent runs with different random seeds) confirms that learned attention patterns are a stable property of the model, not an artifact of graph construction. The slight decrease at Low Risk (95.1\%) reflects inherent ambiguity in low severity scenarios where no single factor strongly dominates. We verified that the Inductive HeteroSAGE (using gradient-based attribution) achieves a comparable agreement rate of 94.8\%, confirming that explanation stability is not exclusive to attention-based models but a property of the learned graph representations.

\textit{Node importance by report type and risk level.} Table~\ref{tab:importance} reports the three most important sensor node types per report category and risk level under the Coverage strategy ($N{=}100$ samples per cell). Non-sensor contextual nodes (Weather Alert, Pre-Alert Time/Severity) are marked with $\dagger$.

\begin{table}[t]
\centering
\caption{Mean attention-based importance (\%) of top-3 nodes per (report type, risk level) under Coverage ($N{=}100$). $\dagger$\,=\,non-sensor contextual node.}
\label{tab:importance}
\scriptsize
\setlength{\tabcolsep}{2.5pt}
\begin{tabularx}{\textwidth}{>{\raggedright}p{0.5cm}>{\raggedright}p{1.2cm}XXX}
\toprule
 & \textbf{Report Type} & \textbf{Low Risk} & \textbf{Medium Risk} & \textbf{High Risk} \\
\midrule
\multirow{8}{*}{\rotatebox{90}{\scriptsize Water/Flood}}
& Leaf Cleanup & Humidity 15.7, App.Temp 2.8, Wind 1.6 & Rainfall 72.6, Humidity 14.4, App.Temp 1.5 & Rainfall 70.8, Humidity 10.5, App.Temp 1.3 \\
\cmidrule{2-5}
& Drainage Maint. & Humidity 14.5, App.Temp 4.0, Wind 1.3 & Rainfall 72.7, Humidity 15.2, App.Temp 2.5 & Rainfall 68.5, Alert$^\dagger$ 53.3, PreTime$^\dagger$ 49.9 \\
\cmidrule{2-5}
& Flood Prev. & Humidity 12.9, App.Temp 4.9, Wind 1.6 & Rainfall 72.7, Humidity 13.4, App.Temp 2.5 & Rainfall 69.9, Alert$^\dagger$ 55.5, Humidity 9.2 \\
\cmidrule{2-5}
& Landslide & App.Temp 6.7, PM 2.0, Temp 2.0 & Rainfall 74.4, Humidity 25.0, Wind 0.1 & Rainfall 68.7, Alert$^\dagger$ 52.1, Humidity 15.4 \\
\midrule
\multirow{6}{*}{\rotatebox{90}{\scriptsize Cold/Snow}}
& Road Icing & Rainfall 61.9, Snow 45.5, App.Temp 15.4 & Snow 69.4, Rainfall 64.2, Humidity 25.0 & Snow 66.3, Rainfall 63.8, PreTime$^\dagger$ 35.6 \\
\cmidrule{2-5}
& Heavy Snow & Snow 51.5, Humidity 16.7, App.Temp 4.2 & Alert$^\dagger$ 65.4, Snow 51.4, Humidity 19.0 & Snow 68.1, PreTime$^\dagger$ 48.9, Sev.$^\dagger$ 18.7 \\
\cmidrule{2-5}
& Cold Wave & Snow 47.4, Humidity 19.9, PM 5.7 & Snow 66.6, Humidity 18.3, PM 5.3 & Snow 66.0, Humidity 15.4, PM 4.3 \\
\midrule
Heat & Heat Wave & Rainfall 60.0, Humidity 14.4, App.Temp 9.3 & Rainfall 64.1, Humidity 16.3, App.Temp 3.5 & Rainfall 57.7, Humidity 14.9, PM 1.0 \\
\midrule
\multirow{6}{*}{\rotatebox{90}{\scriptsize Air Quality}}
& Fine Dust & PM 36.2, Humidity 22.5, Wind 18.0 & PM 75.7, Humidity 21.1, Wind 1.0 & PM 66.6, Humidity 16.8, Wind 0.2 \\
\cmidrule{2-5}
& Wildfire & PM 60.3, Wind 6.2, App.Temp 2.9 & PM 47.5, Humidity 25.3, Wind 3.9 & PM 38.0, Humidity 16.5, Wind 8.4 \\
\cmidrule{2-5}
& Yellow Dust & PM 37.2, Humidity 20.9, Wind 18.2 & PM 74.3, Humidity 20.8, App.Temp 1.1 & PM 65.3, PreTime$^\dagger$ 49.9, PreAlert$^\dagger$ 25.0 \\
\bottomrule
\end{tabularx}
\end{table}

\subsection{Graph Pruning Evaluation}

To evaluate the relevance-driven pruning strategies described in Section~\ref{sec:pruning}, we constructed three graph variants from the \texttt{dataset\_50k\_10\_20} dataset: the \textbf{original} unpruned graph ($\sim$950k edges), the \textbf{bottom-excluded} graph ($\sim$696k edges, $-27\%$), and the \textbf{top-only} graph ($\sim$558k edges, $-41\%$). All three model architectures were retrained from scratch on each graph variant using identical hyperparameters (epochs=50, batch\_size=256, hidden\_dim=256, lr=0.001) on an NVIDIA RTX A6000 GPU. Table~\ref{tab:pruning_comprehensive} presents the complete results across all nine (model, graph) combinations, covering classification accuracy, high risk false positive (FP) and false negative (FN) rates, and training time.

\begin{table}[t]
\centering
\caption{Pruning effect comparison across all model--graph combinations. Best accuracy per model in \textbf{bold}. $\dagger$ = FN $>$ 15\%.}
\label{tab:pruning_comprehensive}
\small
\begin{tabular}{llccccc}
\toprule
\textbf{Model} & \textbf{Graph} & \textbf{Edges} & \textbf{Acc. (\%)} & \textbf{FP (\%)} & \textbf{FN (\%)} & \textbf{Train Time} \\
\midrule
\multirow{3}{*}{Inductive} & Original & $\sim$950k & 84.05 & 18.26 & 4.39 & 33\,s \\
 & Bottom-excl. & $\sim$696k ($-$27\%) & \textbf{87.53} & 9.67 & 9.36 & 34\,s \\
 & Top-only & $\sim$558k ($-$41\%) & 84.74 & 3.82 & 18.81$^\dagger$ & 33\,s \\
\midrule
\multirow{3}{*}{Attention} & Original & $\sim$950k & 86.35 & 9.96 & 10.45 & 3h\,8m \\
 & Bottom-excl. & $\sim$696k ($-$27\%) & \textbf{88.74} & 8.02 & 9.37 & 2h\,47m ($-$11.5\%) \\
 & Top-only & $\sim$558k ($-$41\%) & 82.03 & 15.62 & 12.07 & 2h\,45m ($-$12.2\%) \\
\midrule
\multirow{3}{*}{Multihead} & Original & $\sim$950k & 79.73 & 17.07 & 11.52 & 13h\,4m$^*$ \\
 & Bottom-excl. & $\sim$696k ($-$27\%) & \textbf{85.87} & 15.94 & 8.18 & 7h\,20m ($-$43.9\%) \\
 & Top-only & $\sim$558k ($-$41\%) & 78.58 & 17.37 & 17.10$^\dagger$ & 7h\,29m ($-$42.7\%) \\
\bottomrule
\end{tabular}
\end{table}

\noindent $^*$Multihead original was trained on Apple Silicon (MPS) rather than A6000; direct time comparison is approximate. FP and FN denote the false positive rate (1$-$precision) and false negative rate (1$-$recall) for the high risk class, respectively. Note that absolute accuracy differences across models reflect the specific training constraints of this pruning benchmark (fixed hyperparameters, heterogeneous hardware) rather than a controlled architecture comparison; the table is designed to evaluate the relative effect of pruning within each architecture.

\textit{Accuracy and error rate analysis.}
The bottom-excluded strategy improves accuracy across all three models, with gains ranging from +2.39\% (Attention) to +6.14\% (Multihead). This counterintuitive result--that removing 27\% of edges \textit{improves} performance--demonstrates that the removed edges were not merely uninformative but actively harmful, introducing noise into the message passing aggregation. In heterogeneous graphs with 16 node types and 18 edge types, many structural connections exist between nodes that share no meaningful statistical relationship for a given incident category. Removing these spurious edges sharpens the signal-to-noise ratio at each aggregation step. Notably, the Multihead Attention model benefits most from pruning (+6.14\%), suggesting that architectures with greater representational capacity are more susceptible to noise from irrelevant edges and consequently gain the most from relevance-driven pruning. The bottom-excluded strategy also consistently reduces both FP and FN rates or maintains them within acceptable bounds.

By contrast, the aggressive top-only strategy degrades Attention ($-$4.32\%) and Multihead ($-$1.15\%) accuracy, and exhibits a critical failure mode in safety-sensitive contexts. The Inductive + top-only combination reduces FP dramatically (18.3\% $\to$ 3.8\%) but quadruples FN (4.4\% $\to$ 18.8\%), representing a dangerous trade-off--nearly one in five high risk incidents would be missed. This indicates that mid-ranked nodes (approximately 4th through 10th in importance) carry genuine predictive value; removing them discards meaningful secondary correlations that the model uses for nuanced risk assessment, particularly at the low risk level where multiple weak signals must be combined.

\textit{Graph size reduction and edge efficiency.}
The bottom-excluded strategy removes 27\% of edges by eliminating nodes that the GNN Explainer identified as consistently uninformative across all three risk levels. Despite this substantial reduction (from approximately 950k to 696k edges), every model variant achieves higher accuracy on the pruned graph. The top-only strategy removes a larger proportion ($-$41\%), but the additional 14\% compression beyond bottom-excluded yields diminishing or negative returns--confirming that aggressive signal concentration is counterproductive.

\textit{Training time reduction.}
The computational savings from graph pruning scale with model complexity. The lightweight Inductive HeteroSAGE trains in approximately 33 seconds regardless of graph size, as its bottleneck is the small number of learnable parameters rather than message passing volume. By contrast, attention-based models exhibit substantial speedups because attention weight computation scales with the number of neighbor edges at each aggregation layer. The Attention HeteroSAGE reduces training time from 3h\,8m to 2h\,47m ($-$11.5\%), while the Multihead model benefits most dramatically: 13h\,4m to 7h\,20m ($-$43.9\%). This nearly 6-hour reduction per training run enables more frequent model retraining, faster hyperparameter search, and reduced GPU energy consumption.

Notably, both bottom-excluded and top-only achieve similar training times for a given model (e.g., 2h\,47m vs.\ 2h\,45m for Attention; 7h\,20m vs.\ 7h\,29m for Multihead). This convergence suggests that the training-time bottleneck is dominated by the removal of the first 27\% of low-value edges rather than by further compression.

\textit{Inference time and deployment implications.}
For real-time emergency triage, inference latency is equally critical. Each GNN inference involves two rounds of message passing (2 SAGEConv layers), where every node aggregates features from its neighbors. With 27\% fewer edges, each aggregation step processes proportionally fewer neighbors, reducing per-sample inference time. Combined with the latencies reported in Table~\ref{tab:inference_time} (57--60\,ms for pre-computed anchor strategies), preliminary measurements on the pruned graph show per-sample latency of 43--47\,ms for the Coverage strategy (compared to 57--60\,ms on the original graph), well within the requirements of real-time incident triage systems.

The deployment advantages compound across scale. A metropolitan 311 system processing 500,000 annual reports (approximately 1,370 per day) benefits from three multiplicative effects: (i) reduced graph construction time per report due to fewer node types requiring data retrieval; (ii) faster GNN inference due to fewer message passing edges; and (iii) lower memory footprint enabling deployment on edge devices or lower cost GPU instances. For organizations with budget constraints, the bottom-excluded strategy allows the use of smaller GPU instances without sacrificing accuracy.

The best overall configuration is \textbf{Attention HeteroSAGE + bottom-excluded}: 88.74\% accuracy, 8.02\% FP, 9.37\% FN, with 27\% fewer edges and 11.5\% faster training. The bottom-excluded strategy achieves a rare outcome in graph neural network optimization: simultaneous improvement in accuracy, error balance, training efficiency, and inference speed. This is possible because the pruning is \textit{explanation driven}--it leverages the GNN Explainer's domain learned attribution scores rather than arbitrary structural heuristics, ensuring that only genuinely uninformative connections are removed.

\subsection{Analysis and Discussion}

\textit{Domain appropriateness of primary sensors.} The results demonstrate strong alignment between model attention and physical domain knowledge. Water/flood incidents are consistently dominated by Rainfall at medium and high risk (68.5--74.4\%), with Humidity as a secondary factor. Cold/snow incidents show Snowfall as the primary driver (47--69\%). Air quality incidents are dominated by Particulate Matter (36--76\%). These attributions emerged entirely from the model's learning process without handcrafted feature engineering.

\textit{Heat Wave anomaly.} The Heat Wave category presents a counterintuitive finding: Rainfall consistently ranks first (58--64\%) across all risk levels. This reflects a learned negative correlation--low-rainfall periods coincide with prolonged heat events in the training data, enabling the model to use rainfall as a proxy for heat persistence. Humidity serves as a consistent secondary factor, aligning with heat-index formulations.

\textit{Risk level transition pattern.} A key structural finding is the \textit{low risk ambiguity transition}: at Low Risk, attention is distributed across multiple sensors with no single dominant factor (maximum importance rarely exceeds 20\%), whereas at Medium and High Risk, one or two sensors capture 65--75\% of total attention. This indicates the model has learned to shift from background weather monitoring to hazard specific signal detection as severity increases.

\textit{Emergence of alert nodes at high risk.} Weather Alert and Pre-Alert Time nodes emerge as high importance signals at High Risk for five report categories: Flood Prevention (Alert 55.5\%), Landslide Risk (Alert 52.1\%), Drainage Maintenance (Alert 53.3\%, Pre-Alert Time 49.9\%), Heavy Snow (Pre-Alert Time 48.9\%), and Yellow Dust (Pre-Alert Time 49.9\%). For Heavy Snow at Medium Risk, Weather Alert (65.4\%) even surpasses Snowfall, confirming that the model appropriately integrates structured meteorological warnings alongside raw sensor data.

\section{Conclusion and Future Work}
\label{sec:conclusion}

This paper presented the HA-HeteroGNN framework for assessing the urgency of non-emergency incident reports through the integration of diverse environmental data sources in a heterogeneous graph structure. Our key contributions and findings are summarized as follows.

\textit{Explainability and domain consistency.} The GNN Explainer achieves 97.5\% top-1 node agreement across four structurally different anchor strategies, demonstrating that the model's attribution is robust to input perturbation. The attributed sensor types align with physical domain knowledge: Rainfall dominates flood related incidents (68--74\%), Snowfall drives cold weather events (47--69\%), and Particulate Matter governs air quality categories (36--76\%). These attributions emerge entirely from the learning process without handcrafted feature engineering. Additionally, Weather Alert and Pre-Alert Time nodes emerge as discriminative signals at high risk for 5 of 11 report categories, confirming that the model integrates structured meteorological warnings alongside raw sensor data.

\textit{Explanation-driven graph pruning.} A central contribution of this work is the demonstration that GNN Explainer attributions can be repurposed as a principled graph pruning criterion. The bottom-excluded strategy--removing only nodes identified as consistently uninformative across all risk levels--yields a 27\% reduction in graph edges while \textit{simultaneously improving} classification accuracy across all three model architectures (+2.4\% to +6.1\%). This outcome challenges the conventional assumption that graph pruning necessarily trades accuracy for efficiency. The improvement arises because the removed edges actively introduce noise into message passing aggregation; their elimination sharpens the signal-to-noise ratio at each GNN layer. In contrast, the aggressive top-only strategy ($-$41\% edges) degrades accuracy for two of three models and produces dangerously elevated false negative rates (up to 18.8\%) in safety-critical high risk classification.

\textit{Computational efficiency gains.} The pruned graph delivers substantial reductions in training time that scale with model complexity: 11.5\% faster for the Attention model (3h\,8m $\to$ 2h\,47m) and 43.9\% faster for the Multihead model (13h\,4m $\to$ 7h\,20m). These savings enable more frequent model retraining as new incident data accumulates and reduce GPU energy consumption during deployment. Inference-time message passing also benefits proportionally, with fewer neighbor aggregation operations per sample, enabling sub-50\,ms prediction latency suitable for real-time emergency triage. The best overall configuration--Attention HeteroSAGE with bottom-excluded pruning--achieves 88.74\% accuracy with balanced error characteristics (FP 8.02\%, FN 9.37\%) while requiring 27\% less computation than the unpruned baseline.

\textit{Real-time deployment feasibility.} Three of four anchor strategies achieve per-sample inference latency of approximately 58--60\,ms, confirming that the framework can process incoming reports within the tight time constraints of operational 311 and Safety e-Report systems. Combined with the pruned graph's reduced memory footprint, the system can be deployed on lower cost GPU instances or edge devices, broadening accessibility for municipalities with limited computational budgets.

The causal analysis capabilities demonstrated by our GNN Explainer open a promising direction toward \textit{agentic AI} systems for emergency management. Consider a scenario where a new flood prevention report arrives during heavy rainfall. The GNN Explainer identifies Rainfall and Weather Alert as the dominant influence factors. An agentic AI system could then use this attribution to \textit{plan} which external data sources to query next--real-time precipitation radar, upstream river gauge readings, and active alert bulletins--rather than retrieving all possible data indiscriminately. The system would generate an urgency assessment with supporting evidence and route it to the appropriate response department. Over time, the system could learn which data retrieval plans are most effective for each report category and risk level, creating a feedback loop between explainability and action planning. This integration of explainability driven planning with autonomous data retrieval represents a natural evolution from passive classification to proactive situational awareness.

Several avenues for future research remain. First, the explanation driven pruning methodology should be evaluated on larger-scale real-world datasets to characterize its behavior under distribution shift and class imbalance beyond the current synthetic setting. Second, the synthetic data generation pipeline should be validated against real-world incident data from operational Safety e-Report and 311 systems. Third, the hierarchical attention mechanism could be extended to accommodate additional modalities such as satellite imagery, social media signals, and IoT sensor networks. Fourth, counterfactual perturbation tests--systematically masking or replacing high-attribution nodes and measuring the resulting change in model output--would provide causal validation of the pruning decisions beyond the current correlational evidence. Fifth, adaptive pruning strategies--where the pruning threshold is dynamically adjusted based on incoming data characteristics--could further optimize the accuracy-efficiency trade-off. Finally, the integration with agentic AI frameworks--where the GNN Explainer guides autonomous data retrieval and decision support--represents a high impact direction for operational deployment.


\clearpage
\appendix

\section{Alert-Weather Constraints and Report Classifications}
\label{app:constraints}

Table~\ref{tab:alert_constraints} lists the 12 alert-weather constraints. Table~\ref{tab:report_classifications} summarizes the 11 report categories.

\begin{table}[h!]
\centering
\caption{Alert-weather constraint rules for synthetic data generation.}
\label{tab:alert_constraints}
\scriptsize
\begin{tabular}{llcc}
\toprule
\textbf{Alert Type} & \textbf{Sensor} & \textbf{Advisory Range} & \textbf{Warning Range} \\
\midrule
Heavy Rain & Rainfall (mm) & 60--90 & 90--150 \\
Heat Wave & Temperature ($^\circ$C) & 30--35 & 35--40 \\
Cold Wave & Temperature ($^\circ$C) & $-20$ to $-10$ & $-25$ to $-15$ \\
Heavy Snow & Snowfall (cm) & 20--50 & 50--100 \\
Yellow Dust & PM$_{2.5}$ ($\mu$g/m$^3$) & 200--400 & 400--800 \\
Typhoon & Wind Speed (m/s) & 20--35 & 35--50 \\
\bottomrule
\end{tabular}
\end{table}

\begin{table}[t]
\centering
\caption{The 11 emergency report classifications and seasonal assignments.}
\label{tab:report_classifications}
\scriptsize
\begin{tabular}{lll}
\toprule
\textbf{Season} & \textbf{Report Type} & \textbf{Primary Sensor} \\
\midrule
\multirow{3}{*}{Spring} & Yellow Dust Report & PM$_{2.5}$ \\
 & Fine Dust Report & PM$_{2.5}$ \\
 & Wildfire Prevention & Humidity, Wind \\
\midrule
\multirow{4}{*}{Summer} & Drainage Maint. & Rainfall \\
 & Landslide Risk & Rainfall \\
 & Heat Wave & Temperature \\
 & Flood Prevention & Rainfall \\
\midrule
Autumn & Leaf Cleanup & Rainfall \\
\midrule
\multirow{3}{*}{Winter} & Heavy Snow & Snowfall \\
 & Road Icing Prev. & Snowfall, Temp \\
 & Cold Wave & Snowfall, Temp \\
\bottomrule
\end{tabular}
\end{table}

\section{Node Type Feature Specifications}
\label{app:node_features}

\begin{table}[h!]
\centering
\caption{Complete node type specifications for the heterogeneous graph.}
\label{tab:app_node_features}
\scriptsize
\begin{tabular}{lllc}
\toprule
\textbf{Category} & \textbf{Node Type} & \textbf{Encoding} & \textbf{Dim} \\
\midrule
\multirow{7}{*}{Sensor} & \texttt{sensor\_rainfall} & 5-bin categorical + norm & 6 \\
& \texttt{sensor\_temperature} & 5-bin categorical + norm & 6 \\
& \texttt{sensor\_humidity} & 5-bin categorical + norm & 6 \\
& \texttt{sensor\_apparent\_temp} & 5-bin categorical + norm & 6 \\
& \texttt{sensor\_wind\_speed} & 5-bin categorical + norm & 6 \\
& \texttt{sensor\_pm} & 5-bin categorical + norm & 6 \\
& \texttt{sensor\_snowfall} & 5-bin categorical + norm & 6 \\
\midrule
\multirow{4}{*}{Alert} & \texttt{weather\_alert} & One-hot (13 types) & 13 \\
& \texttt{pre\_alert\_type} & One-hot (13 types) & 13 \\
& \texttt{pre\_alert\_time} & One-hot (6 bins) & 6 \\
& \texttt{pre\_alert\_severity} & One-hot (3 levels) & 2 \\
\midrule
\multirow{2}{*}{Context} & \texttt{drainage} & 3-category one-hot & 3 \\
& \texttt{location} & Lat/lon + region + dist & 4 \\
\midrule
\multirow{3}{*}{Report} & \texttt{report} & Report ID & 1 \\
& \texttt{report\_type} & Semantic embedding & 8 \\
& \texttt{report\_count} & 5-bin categorical & 5 \\
\bottomrule
\end{tabular}
\end{table}

\end{document}